\documentclass[pmlr]{jmlr}

\RequirePackage{graphicx}
 \usepackage{booktabs}
\usepackage{longtable}
\makeatletter
\def\set@curr@file#1{\def\@curr@file{#1}} 
\makeatother
\usepackage[load-configurations=version-1]{siunitx} 

\theorembodyfont{\upshape}
\theoremheaderfont{\scshape}
\theorempostheader{:}
\theoremsep{\newline}

\title[Medical Knowledge Graph Automation (M-KGA)]{Accelerating Medical Knowledge Discovery through Automated Knowledge Graph Generation and Enrichment}

\author{\Name{Mutahira Khalid}
       \Email{Mutahira.Khalid@tib.eu}\\ 
       \addr TIB – Leibniz Information Centre for Science and Technology\\
       Hannover, Lower Saxony, Germany 
       \AND
       \Name{Raihana Rahman}
       \Email{rrahman@college.Harvard.edu}\\ 
       \addr Department of Statistics\\
       Harvard University\\
       Cambridge, Massachusetts, United States
       \AND
       \Name{Asim Abbas}
       \Email{axa2233@student.bham.ac.uk}\\ 
       \addr School of Computer Science\\
       University of Birmingham,\\
       Edgbaston, B15 2TT, Birmingham, England, UK
       \AND
       \Name{Sushama Kumari}
       \Email{sushamawork@gmail.com}\\ 
       \addr Sr Technical Program Manager\\
       Routing and Planning\\
       Amazon\\
       Bellevue, Washington, United States
       \AND
       \Name{Iram Wajahat}
      \Email{wajahati@stjohns.edu}\\ 
       \addr Institute for Biotechnology\\
       St. John's University\\
       New York, New York, United States
       \AND
       \Name{Syed Ahmad Chan Bukhari*}
       \Email{bukharis@stjohns.edu}\\ 
       \addr Division of Computer Science, Mathematics and Science\\
       St. John's University\\
       New York, New York, United States\\
       *Correspondence should be directed to: bukharis@stjohns.edu}

\begin{document}

\maketitle

\begin{abstract}
  Knowledge graphs (KGs) serve as powerful tools for organizing and representing structured knowledge. While their utility is widely recognized, challenges persist in their automation and completeness. Despite efforts in automation and the utilization of expert-created ontologies, gaps in connectivity remain prevalent within KGs. In response to these challenges, we propose an innovative approach termed ``Medical Knowledge Graph Automation (M-KGA)". M-KGA leverages user-provided medical concepts and enriches them semantically using BioPortal ontologies, thereby enhancing the completeness of knowledge graphs through the integration of pre-trained embeddings. Our approach introduces two distinct methodologies for uncovering hidden connections within the knowledge graph: a cluster-based approach and a node-based approach. Through rigorous testing involving 100 frequently occurring medical concepts in Electronic Health Records (EHRs), our M-KGA framework demonstrates promising results, indicating its potential to address the limitations of existing knowledge graph automation techniques.
\end{abstract}

\section{Introduction}
While once considered a relic of early Artificial Intelligence (AI) research \cite{smolensky1987connectionist}, knowledge graphs (KGs) have experienced a remarkable resurgence in recent years. Knowledge graphs, which serve as the foundation of symbolic AI, consist of interconnected knowledge pertaining to many domains like medical, finance, commerce, and education \cite{zou2020survey}. Particularly in medicine, KGs have emerged as indispensable tool.

KGs offer numerous advantages over traditional relational databases, primarily stemming from their diverse array of nodes and the ability to establish connections between them. This versatility lends itself to applications ranging from search engine optimization to recommendation systems, knowledge discovery, and research facilitation. However, the process of constructing KGs is inherently labor-intensive, especially in the intricate domain of medicine, despite its profound significance.

A KG is essentially a multi-graph, characterized by its directed, labeled, and diverse nature. At its core, a KG comprises facts, often represented as triplets \cite{hogan2021knowledge}, each consisting of a relationship and two nodes. With KGs consisting of millions to billions of these triplets, their aggregation holds immense potential for information discovery, data integration, and effective management. Yet, crafting KGs, particularly within the medical domain, presents formidable challenges due to the complexity of medical concepts and relationships between them. Compounding these challenges is the prevalence of unstructured medical data, further complicating the KG creation process.

Various methodologies for graph creation have emerged in recent years, ranging from automatic to semi-automated and manual approaches \cite{hao2021construction}. While these methods addresses some challenges. However they often suffer from significant deficiencies such as they lack standardized platforms or code for graph creation despite offering graphical methodologies. Furthermore, some approaches utilize hospital notes to generate nodes and relations, they overlook the potential benefits of data augmentation, resulting in incomplete graphs. Additionally, there is currently no promising technology capable of generating graphs in real-time, further impeding the process.

\begin{figure}
    \centering
    \includegraphics[width=6in]{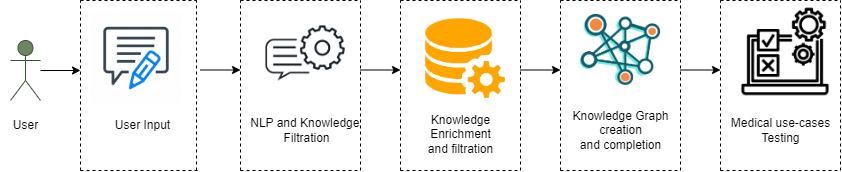}
    \caption{A flow diagram of the proposed M-KGA framework.}
    \label{fig:ds}
\end{figure}

In response to these challenges, our proposed approach, Medical Knowledge Graph Automation (M-KGA), effectively addresses these obstacles by seamlessly processing both structured and unstructured data in real-time. The preprocessing steps involve Named Entity Recognition (NER)-based keyword extraction from unstructured data using the SciSpacy library \cite{neumann2019scispacy}, tailored for scientific and biomedical content. Subsequently, a knowledge filtration phase eliminates duplicates and extraneous terms before rapidly generating the knowledge graph using Neo4j's query language, `Cypher'. Furthermore, we leverage Bioportal \cite{noy2009bioportal} for data augmentation, enriching medical terms semantically via incorporating metadata such as definitions, synonyms, and hierarchies. Following data augmentation, a semantic information filtration phase removes duplicates and non-English terms, enhancing the quality of the knowledge graph.

To uncover concealed linkages and associations between medical terms, we utilize the pre-trained contextual word embedding model Clinical BERT \cite{alsentzer2019publicly}, trained on the MIMIC-III dataset. This facilitates the discovery of valuable insights within the data and contributes to the creation of a comprehensive knowledge graph. We proposed cluster-based and node-based comparison methods to unveil hidden relationships by exploiting Clinical BERT within the knowledge graph.

Furthermore, our proposed approach enables users to effortlessly navigate complex features and generate autonomous knowledge graphs. Consequently, users can efficiently generate KGs based on input data, eliminating the need for prolonged waiting periods. Additionally, users have access to the generated files for further study and analysis. Ultimately, the discovery of hidden connections through our approach aids clinicians in gaining a deeper understanding of patient symptoms. Also it is benefiting insurance companies in identifying fraudulent claims and examining inaccurate forecasts of medical codes.

\subsection*{Generalizable Insights about Machine Learning in the Context of Healthcare}
In summary, our research makes the following contributions:
\begin{enumerate}
\item Proposed a significant approach for automating the construction of a Medical Knowledge Graph, known as Medical Knowledge Graph Automation (M-KGA).
\item Utilization of node-based and cluster-based comparisons for KG completion.
\item Conducting rigorous evaluation to demonstrate the efficiency of our technique and the resulting knowledge graph.
\end{enumerate}

Further, the paper is arranged as: Related work on KG automation in the medical and other fields is discussed in Section II. The proposed methodology to create KG is presented in Section III. The result and assessment is describe in Section IV, conducted on several medical use cases. Finally, we have comprehensively presented the limitation, future direction and conclusion in Section V.

\section{Related Work}

Over the past few years, a number of knowledge graph automation techniques have emerged in every aspect of life, including business, healthcare, finance, and education. Most of them were proposed with particular use cases and problems in mind. Given its diversity, volume, velocity, and truthfulness, the KG creation in the healthcare industry is far more challenging than in other industries. Furthermore, KG completion approaches are seldom utilized despite being a valuable tool for finding relevant connections through the use of machine learning to evaluate large amounts of data. Aside from these restrictions, no platform exists that can be used to offer an automated full KG for any subfield in the medical industry.

\begin{table*}[t]

    \label{crouch}
    \begin{tabular}{|p{0.10\linewidth}|p{0.20\linewidth}|p{0.15\linewidth}|p{0.20\linewidth}|p{0.225\linewidth}|}

\hline
Related Work & Domain & Dataset & Strengths & Limitations \\ \hline
\cite{midha2021towards} & Domain-independent statistical data & open Nova Scotia & Domain-Independent & Missing KG completion \\ \hline
\cite{malik2020automated}& Subarachnoid hemorrhage & 1000+ patient records & Detailed methodology, Open-source code & Missing KG completion, difficult to generalize to other medical diseases/ areas\\ \hline
\cite{sahlab2022knowledge} & Domain-independent & Research Papers & Linking research  & Better scholary Graphs are present \\ \hline
\cite{oelen2020creating}&Systematic Literature review papers & Research papers & Knowledge Retrieval,
open-source
 & Semi-automated approach \\ \hline
\cite{alam2023automated}& Medical domain & Unstructured patient records & Automated Approach & Restricted to only Covid-19 and cerebral aneurysm use-case \\ \hline
\cite{sun2020medical} & Medical claims & Medical texts and Chinese Food and Drug Administration & Fraud claim detection & Require human intervention \\ \hline
\cite{li2020real} & Healthcare & Patient records & KG completion & Semi-automated approach \\ \hline
\cite{zhang2023saka}& User provided data, healthcare & run-time data input & KG construction and audio data, Knowledge management & Semi-automated approach, unavailable open-source code\\ \hline
\cite{wang2020covid} & Medical & Covid-19 data, scientific literature (140k papers) & High quality KG & Semi-automated apporach\\ \hline

        \bottomrule
        
    \end{tabular}
    \caption{Latest Research on Knowledge Graph Automation Approaches.}
\end{table*}

Some intriguing use-cases for graph automation systems include systematic literature reviews \cite{sahlab2022knowledge}, in which research article data is gathered from electronic databases like IEEExplore, ACM, Springer Link, Wiley, and ScienceDirect. Using this data, a knowledge graph was constructed that can aid in systematic literature review by offering better article recommendations. Although the concept of linking research data is sound, the approach given does not appear to yield particularly effective results, as it might be challenging to identify a paper's research focus and substance. Additionally, there exist more effective knowledge graphs for this purpose, such as the Hi-knowledge graph, the Microsoft Academic Graph, the Papers with Code graph, the Cooperation Databank, etc. An analogous approach involved extracting information from comprehensive literature reviews and incorporating it into the Open Research Knowledge Graph platform as a knowledge graph \cite{oelen2020creating}. Tables were employed as the data source in this semi-automated method, which produced several triplets from a single table row. These created graphs are not just useful in the medical field; they may also help with fact discoveries in various domains.

Medical KG are built using a variety of approaches, including human, semi-automatic, and automated methods, as well as modern and conventional procedures. An important example of a semi-automated graph generation procedure that made use of scientific literature and pre-existing datasets is the COVID-19 graph \cite{wang2020covid}. Evidence mining, hypothesis ranking, and relation extraction were carried out using hierarchical spherical embeddings, ontology-enriched text embeddings, and cross-media semantic-structure representation. A mechanism for producing reports and responding to inquiries was also devised. 7,230 diseases, 9,123 chemicals, and 50,864 genes are included in the final KG. There are 1,725,518 chemical-gene relationships, 5,556,670 chemical-disease ties, and 77,844,574 gene-disease links.

A four-step process for creating knowledge graphs semi-automatically from statistical data was described \cite{midha2021towards}. These processes include data collection/acquisition, knowledge extraction, knowledge fusion, and knowledge storage. Using DOID, Geonames, DBpedia, and other resources, an ontology for publicly available statistical datasets was created in addition to the graph creation. The data was not restricted to any one domain, and a great deal of connecting was done to produce final knowledge that could be applied to advanced data analytics. The research presented by the authors is innovative, but since they are utilizing data from several sectors, there may be countless other avenues for investigation. Although they haven't employed KG Meddings models for fact discovery, the statistical data can yield some fascinating discoveries. Since hundreds of others will be able to examine the relevant data and obtain graphs for use in their own professions, the researchers ought to make this available as open-source code.

Another method for creating healthcare knowledge graphs was available; it consisted of eight modules and produced a KG semi-automatically utilizing 16,217,270 de-identified clinical visit data from 3,767,198 patients \cite{li2020real}. Entity recognition, entity normalization, relation extraction, property calculation, graph cleaning, related-entity ranking, and graph embedding are among the processes in the process. To store extra context in healthcare graphs, the quadruplet form is used instead of the traditional triplet. A medical KG of 22,508 entities and 579,094 quadruplets with nine distinct entity kinds was the end result.

Knowledge graphs are useful in many different contexts, including medicine, diagnosis, and information acquisition. Another crucial use case is the detection of fraudulent claims; insurance firms employ several medical coders to confirm claims. A Fraud, Waste and Abuse (FWA) detection system was developed to identify fraudulent claims using a Chinese medical knowledge graph \cite{sun2020medical}. The graph was created in a semi-automated fashion, with human interaction being done to confirm the graph's legitimacy after entities were extracted using deep learning-based techniques. Information about diseases was taken from medical texts and drug labels were gathered from the Chinese Food and Drug Administration. With 1,616,549 nodes and 5,963,444 relations, the final graph had a 70\% accuracy rate in detecting fraud claims.

KGs are frequently designed for text data because it makes up nearly 80\% of all available data. Semi-automated KG Construction and Application (SAKA) \cite{zhang2023saka} is an intriguing use case for KG generation that was created. It uses both auditory and structured data to create graphs. Voice Activity Detection (VAD), Speaker Diarization (SD), and the Medical Information Extractor (MIE) model are the methods used in the audio-based KG Information Extraction (AGIE) technique to extract entities. In addition to building graphs, a system was created to receive user inquiries and extract pertinent data. The knowledge management module continuously verifies that the data is current and relevant. Testing the method on the LibriSpeech, VoxCeleb, and doctor-patient dialogue datasets yielded a respectable performance.

An instance of a medical knowledge graph for a certain medical field: “Subarachnoid hemorrhage” involved the use of over a thousand case records \cite{malik2020automated}. The medical entities were enhanced by utilizing the appropriate Bioportal ontologies. Multiple layers, including the semantic knowledge, statistical knowledge layer, predictive knowledge layer, and knowledge factory layer, were involved in the graph generation process. Though it's tough to use the source code for other medical fields, it is available to the public. Although it might seem like switching ontologies would make it feasible in other contexts, research indicates that this is not so simple, since word embeddings for subarachnoid use cases need to be taught among other things. The KG completion job, which can aid in creating linkages for many findings, was also overlooked by the authors in this case. A different automation method was created for use cases including evidence-based medicine (Alam, 2023). With peer-reviewed ontologies, the KG was automatically built for cerebral aneurysm and COVID-19. To create and finish the graph, machine learning based clustering models were utilized. Also deep learning techniques such  as RNN, BioBERT, etc were used. The results demonstrated 93\% and 82\% accuracy on the COVID and aneurysm data sets, respectively.

The majority of approaches depict the KG creation process in a manual or semi-automated manner, as the literature demonstrates. The majority of them are tailored to specific use cases and the medical field, and they rarely incorporate the KG completion technique, which can harness the potential of BIG data to uncover facts. Although the generated KGs are indeed helpful in many ways, they cannot be applied generally. Furthermore, there isn't a platform that can handle user requests to validate specific use cases and create the appropriate KG in a matter of minutes or seconds. Our method, which offers automation, is distinct and creative in that it uses expert-created ontologies to produce a full and comprehensive KG while satisfying the user's request for a specific KG generation. People can benefit from the approach in a variety of fields, and the KGs produced by our method can be used to enhance the research of others.  

\section{Methods}
Our proposed approach is designed in mutiple steps. Figure 2 illustrates the entire workflow of the Medical-Knowledge Graph Automation (M-KGA). This approach acquire data in two formats: structured and unstructured. It then applies various natural language processing (NLP) techniques to process the data. Initially, Bioportal is utilized to identify and enhance medical concepts with semantic information. The fetched data is filtered and used to create nodes in a knowledge graph (KG) along with their relationships. A pre-trained contextual word embedding model Clinical BERT is leaveraged to discover hidden connections for KG completion. Finally, a Cypher query file is generated to facilitate the creation of the KG in Neo4j. The details of each individual stage are outlined below.

\subsection{User Input}
The M-KGA technique allows users to input medical data in two distinct formats: structured and unstructured. When we say "structured," we mean that the user defines the medical terms with precision. The data does not contain any interconnected notions. Here is an example of a text that is organized in a structured manner: 

\begin{quote}
Structured Input Example:\\
\noindent \textit{ [‘fever’, ‘diarrhea’, ‘insomnia’, 'severe acute respiratory syndrome’, ‘diabetes’]}
\end{quote}

Unstructured text, on the other hand, is free natural language text that is understandable to people but not to computers. This is the text written by a medical professional for instance diagnosis of a patient. An illustration of an unstructured text is:

\begin{quote}
Unstructured Input Example:\\
\noindent \textit{ [``If you have a condition called polyuria, it's because your body makes more pee than normal. Adults usually make about 3 liters of urine per day. But with polyuria, you could make up to 15 liters per day. It's a classic sign of diabetes."]}
\end{quote}

The developed code can take data in both formats; if structured text is needed, it will ask for the data numerous times. If unstructured formatted text is needed, it will accept it all at once and find out the concepts on its own.
\subsection{NLP-based Knowledge Filtration}

We have introduced two approaches for NLP based knowledge filtration consisting of i) NER-based keyword extraction and ii) knowledge filtration. The NER-based keyword extraction attempts to identify medical concepts from unstructured text in the first place, and transform into structure format. Subsequently the Key Knowledge Filtration process is employed to filter the most prominent information obtained from the previous step. Each process is comprehensively presented below:

\subsubsection{NER-based Keyword Extraction:}
An unstructured document is provided as input to extract a list of clinical entities employing NER-based keyword function. This process is integral for populating the nodes of the medical knowledge graph, enabling the incorporation of diverse medical entities such as diseases, treatments, and other clinical concepts. We have leveraged the SciSpacy library \cite{alsentzer2019publicly}, a specialized extension of the popular spaCy natural language processing framework tailored for scientific and biomedical text. Specifically, the function loads the en\_core\_sci\_sm model, which is optimized for processing biomedical text. 

The NER-based Keyword Extraction function requires the unstructured clinical text as input, as a result clinical concept would be extracted as an output. Further, this approach then leverages the biomedical model to process the text and extract clinical entities, returning a list that can be seamlessly integrated into the evolving medical knowledge graph. 

\subsubsection{Knowledge Filtration:}
Subsequently, acquiring a list of clinical or medical concepts, knowledge filtration function would be applied to choose only relevant and prominent concepts. The concepts that are previously extracted may contain some redundancies. Knowledge Filtration looks for duplicates in the data by Fuzzy-matching and filters it further. Additionally, the knowledge filtration assists the medical practice in decision making towards building a comprehensive KG by including and excluding certain concepts. The concept that is extracted during the NER-based keyword extraction process may be not relevant to current problem, disease diagnosis or treatment.  By mapping these words to Bioportal Ontologies, more filtering is applied. 

\begin{figure*}
    \centering
    \includegraphics[width=6.6in]{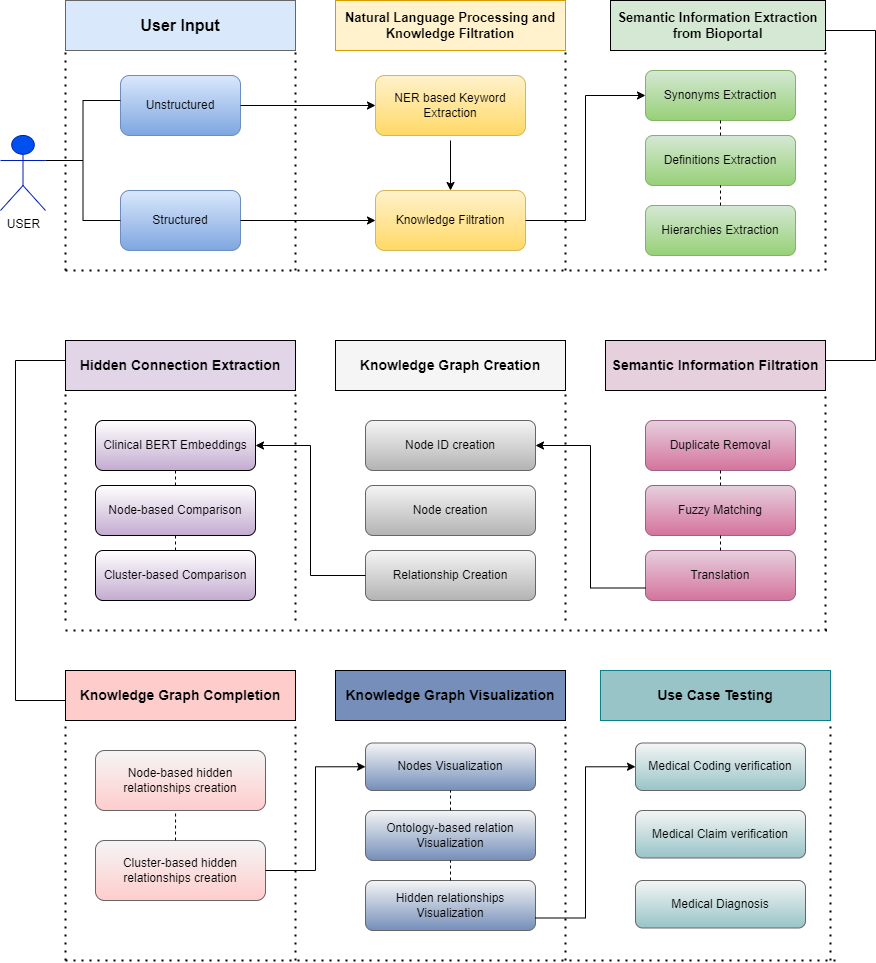}
    \caption{Medical-Knowledge Graph Automation Approach (M-KGA).}
    \label{fig:ds}
\end{figure*}

\subsection{Semantic Information Extraction}
The organized, enhanced, and sifted material from the preceding stage is used in the semantic information extraction step. Mapping these terms to expert-created Bioportal ontologies \cite{noy2009bioportal} allows you to retrieve the semantically enriched information using the Bioportal REST API. Our approach's strength is rooted in the notion that M-KGA is not exclusive to any particular medical condition or illness. Instead, it attempts to enrich data without being limited to particular ontologies.  

Different sorts of semantically enriched information are retrieved from ontologies. Synonyms are words, phrases, or morphemes that share the same meaning as the term being mapped. Definition: A group of terms or phrases that provide a longer description of the term that is mapped. In this process, the two types of extraction took place this is presented below.

\subsubsection{Semantic Knowledge Filtration}
Since our method is not limited to a specific medical condition or issue, M-KGA anticipates a high degree of data diversity, variation, and redundancy. In this step, data that has been semantically enhanced is filtered using a variety of techniques. The method attempts to translate data from many languages to English, eliminate duplicates from the retrieved data, and then use fuzzy-matching to further filter the findings. 

\subsubsection{Translation:}
The data used in the enrichment stage is multilingual and comes from many ontologies. To translate this material, libraries were explored. We need to take this action since, in the absence of translation, we will lose some important information. Text that we identified as non-English was translated into English; if the language cannot be identified or there are any exceptions, the text is eliminated. The subsequent stage does not include this deleted portion. Since the non-English text cannot be used in any further steps, it has been removed. Pre-trained model employed to find hidden connections are unable to comprehend the data, which will result in problems. Aside from that, this step will address the limitations imposed by Neo4j on the creation of a Cypher Node ID.

\subsubsection{Duplicate Removal:}
This step takes the enriched data from the previous stage and tries to remove the duplicates. As we are fetching data from 1000+ Bioportal ontologies, we will likely data massive duplications. In this step we used semantic information, changed all synonyms and definitions to lower case and then used simple set operation on python to remove redundant entries.

\subsubsection{Fuzzy Matching:}
An expansion of the duplicate removal process is fuzzy matching. Certain enriched data may contain semantically comparable text that cannot be removed with set procedures. In an effort to maintain the content's uniqueness, we employed this strategy. This also fixes problems in the ID creation stage and greatly aids in the removal of semantically duplicate items.

\subsection{Knowledge Graph Creation}
Creating a KG is a challenging endeavor in and of itself because it requires extreme caution while creating nodes and interactions. The KG was developed in Neo4j’s Cypher query language. The format has unique limitations. The node ID in Cypher ought to begin with a character rather than a number, special character, non-English phrase, etc. Taking these factors into account, we produced graphs. This will be further explained in the steps following.

\subsubsection{Node ID creation:}
Nodes ID was developed with the understanding that hundreds of connections—both hidden and provided by the ontology—must be made between nodes. We translated the Node content or enriched data into ID by adhering to the ID requirements for different KG formats in order to reduce the amount of computing resources required for ID retrieval for comparison and connection formation. As previously stated, the Node ID in Cpyher only accepts data in English format; special characters are not permitted, etc. Using this method, the enriched data on polyuria is transformed into IDs such as "excessivesecretionofurine," from the definition of "excessive secretion of urine." Therefore, we don't have to go look for the ID connected to that Node every time we need to establish a connection. All we had to do was apply our function and turn the content into ID.

\subsubsection{Node Creation:}
This step builds the nodes for the structured and unstructured (converted to structured) data, as well as the semantically enriched data, according to the ID creation technique previously outlined. Different kinds of nodes have been created. Synonyms, medical concepts, definitions and so on are among the categories. Depending on its kind, every node in the graph is represented by a distinct color. The node displays the content. All KG nodes are constructed in this step.

\subsubsection{Relationship/ Connection creation:}
This step connects different nodes based on the expert-provided/ ontology-provided connections. With each iteration, a semantically enriched node is created with ID, the connection creation step, uses the ID and connects the node with the main medical concept. Same is the case for all semantically enriched data. Here, relationships are also of different types such as synonyms, definitions, and so on. Relationships are labeled and directed.

\subsection{Hidden connection Extraction}
The earlier processes collect user data, filter it, obtain enriched data from ontologies produced by experts, and produce a knowledge graph. In addition to the links supplied by experts, our method looks for hidden connections absent from ontologies. The ontologies offer richer medical terminology, but it might be challenging to determine whether or not these concepts are related to one another. Exist any connections that could be omitted to improve the analysis of the medical data? We made knowledge graphs, but how do we complete them? 

To address these problems, we attempted using KG embeddings for our method, which can predict links given KG triplets. Sadly, these methods are ineffective for tiny graphs. Our method can create both large and tiny KG in response to user requests; nevertheless, KG embeddings are unable to function on small networks because these models need thousands of triplets. Therefore, in order to create connections, we took advantage of word embeddings to determine a word's meaning and relationships with other words. It should be highlighted that our method looks for connections with other medical concepts and their enriched content rather than trying to establish links with its semantic enriched nodes, which are all already connected.

\subsubsection{Clinical BERT Embeddings:}
We took advantage of Clinical BERT embeddings to extract vector representation of medical concepts and their contextual meta-data. We utilized the Clinical BERT embeddings, which are trained on a sizable medical corpus, in place of creating our own model. Medical Information Mart for Intensive Care III (MIMIC-III) is used to train the model.  We took use of their pretrained nature and open-source nature to comprehend medical concepts and their interrelationships. We calculated the degree of similarity between various terms and built relations that led to knowledge graph completion based on the distance and user-defined threshold.

\subsubsection{Cluster-based Comparison:}
We offered two methods for locating the links that are buried in knowledge graphs. We treated every medical concept and its semantically enriched data as a cluster in a cluster-based method. Using all available semantic information, we composed a paragraph and then used the Clinical BERT model to look for embeddings. Clinical BERT implementation is not scalable and introduces mistakes on big clusters. In order to address this method, we segmented the paragraph into chunks, handled exceptions, fetched embeddings for each chunk and then divided by the total number of pieces. A cluster is mapped with other clusters according to a user-defined threshold. The threshold and the degree to which users require specific or general connections are key factors here. In actuality, the threshold is the separation between the clusters. To determine whether clusters have strong relationships or not, users can choose the lower threshold. Cluster-based comparisons or connections lead to KG completion quickly and at low computing cost. This step introduces further relationships named 'embedding\_match\_cluster' in the KG. Figure 3 is an example of Cluster-based comparison method.

\begin{figure*}
    \centering
    \includegraphics[width=7in]{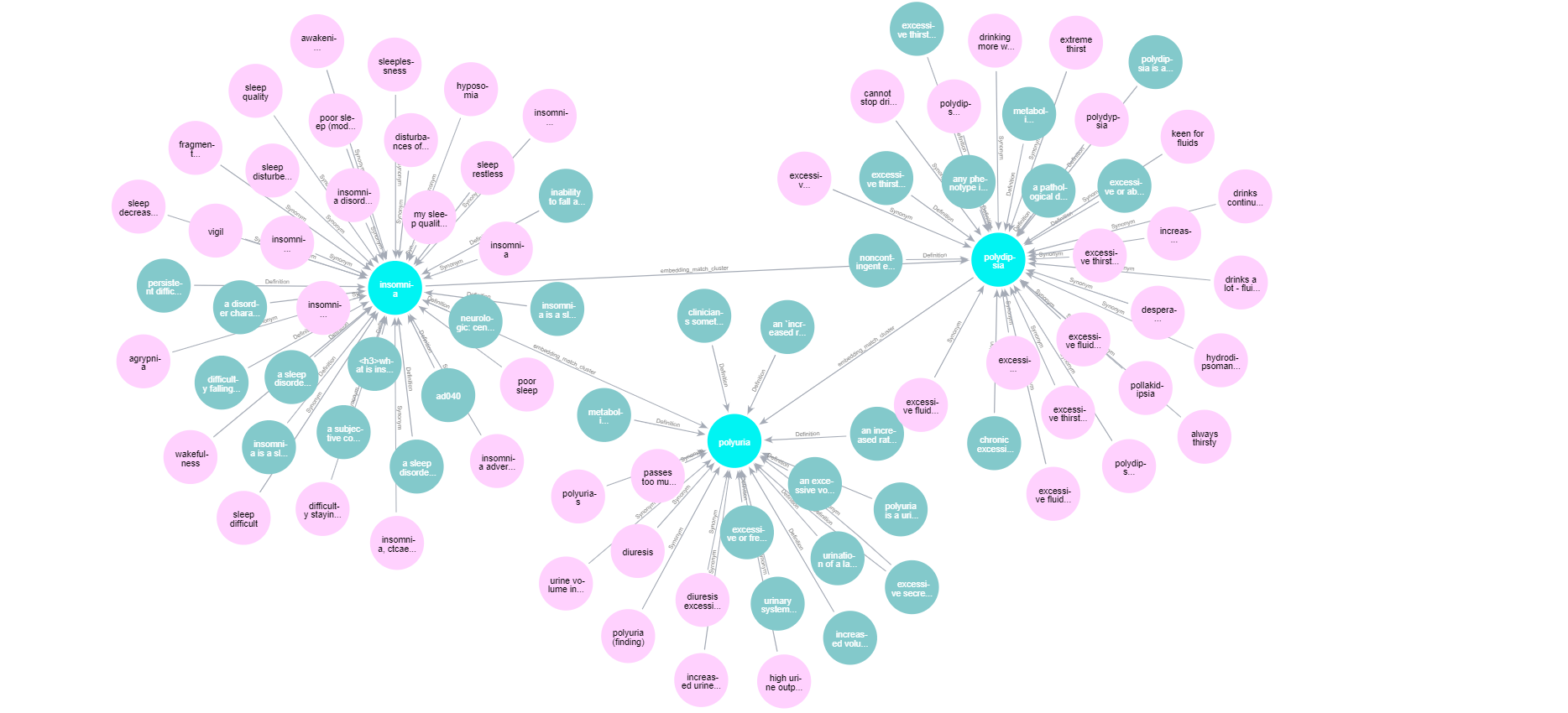}
    \caption{Cluster-based comparison on Medical diagnosis use-case.}
    \label{fig:ds}
\end{figure*}

\subsubsection{Node-based Comparison:}
In contrast to the cluster-based approach, the node-based technique looks for connections with other nodes in the cluster. Using word embeddings, this compares a single node to every other cluster's node based on nodes. The lack of a large amount of text on the nodes means that scalability is not an issue. Furthermore, this method requires around n2 time and is computationally costly, as opposed to the cluster-based method. Because the node-based technique allows us to determine the exact match of the link, it is much easier to understand. Here, connections are also established according to user-specified thresholds. Depending on the size of the graph, connecting nodes takes minutes. Figure 4 shows the result depiction of this method. This adds relationship named 'embedding\_match\_node' in the KG.

\begin{figure*}
    \centering
    \includegraphics[width=6in]{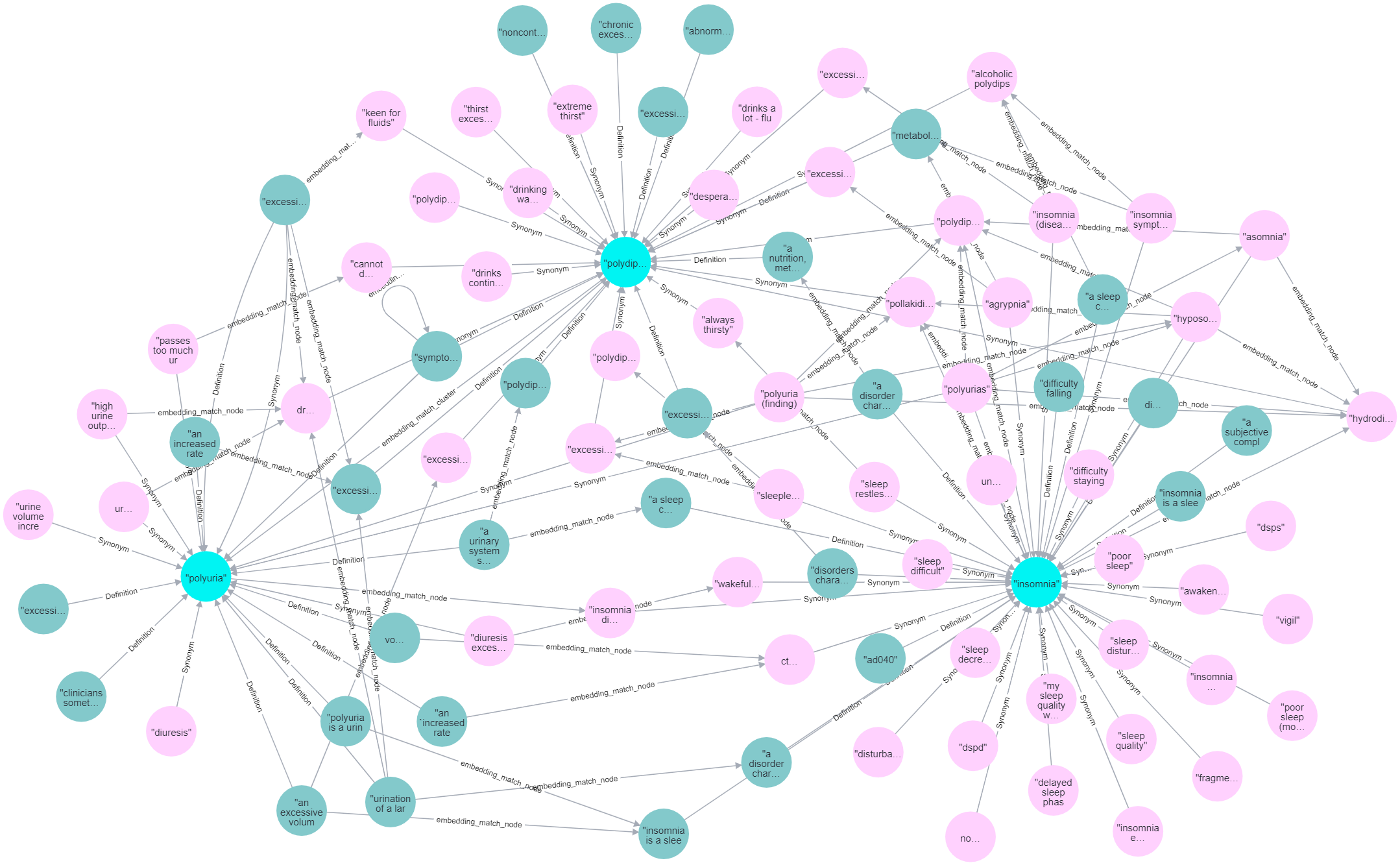}
    \caption{Node-based comparison on Medical diagnosis use-case.}
    \label{fig:ds}
\end{figure*}

\subsection{Use case Testing}
To determine the efficacy of our method, we ran three separate use cases through it. Three use cases were selected: medical claims, medical diagnostics, and medical coding. We used the Spanish dataset known as "CodiEsp" [8] to test each of these use cases.  Experts annotate CodiEsp data with ICD-10 codes. To test our strategy, we used the discharge summaries along with their annotations.

\subsubsection{Medical coding and Claim verification:}
We used several synopses, medical terms, and related medical codes. We extracted the medical terms contained in the descriptions of the medical codes after converting them into descriptions. We then used our Medical-Knowledge Graph Automation (M-KGA) technique to see the outcomes after passing all the medical concepts—such as discharge summaries or descriptions of diagnostic codes—through it. True positive and true negative cases were used in our tests. We took the summaries, applied our method to their annotations, annotated the summaries once again with fictitious examples, and retested the method. The approach demonstrated its usefulness through visualization and proved satisfactory in all cases. We also experimented with various thresholds.

\subsubsection{Medical Diagnosis:}
We also tested the method for medical diagnosis using CodiEsp data. Each medical summary's knowledge graph was made using the concepts that were extracted from the summaries using the NER-based keyword extraction stage. We applied both node-based and cluster-based comparisons, and we generated a complete KG. We presume that any medical ideas included in a summary must be related to one another; this relationship will confirm the usefulness of our method for completing graphs and offer a far more profound comprehension of the relationships taken from ontologies. Using our node-based and cluster-based comparisons technique, the strategy demonstrated significance in all experiments and most medical terms within the same summary generated links. To better understand the approach's operation, we also put it to the test with negative cases as well.

\section{Evaluation}
In this section, we present the results obtained from the implementation and testing of our proposed Medical Knowledge Graph Automation (M-KGA) approach. The evaluation aims to assess the effectiveness and efficiency of M-KGA in constructing comprehensive knowledge graphs from medical concepts provided by users. We conducted experiments using a diverse set of 100 medical concepts to evaluate the performance of our approach across various domains within healthcare.

In our evaluation, we partitioned the 100 medical concepts into two sets: 50 for assessing the cluster-based comparison method and another 50 for evaluating the node-based comparison approach. Each set underwent pairing, facilitated by the GPT-3.5 model, to create pairs of medical concepts. These pairs were compiled into an Excel file for annotation by human medical experts. The experts were tasked with annotating each pair based on measures of True Positive (TP), False Positive (FP), True Negative (TN), and False Negative (FN), providing valuable insights into the accuracy and performance of our approach.

Following the annotation process, we applied both the cluster-based and node-based comparison methods to the pairs of medical concepts with threshold=4. Utilizing these methods, we constructed knowledge graphs for each pair and analyzed whether they successfully identified connections as annotated by the experts. Given that traditional ontologies often struggle to find connections among certain medical concepts, our objective was to determine if our approach could uncover hidden connections that might otherwise remain undiscovered. This analysis aimed to ascertain the efficacy of our proposed methods in augmenting existing knowledge and revealing previously unrecognized relationships within the medical domain.

\begin{figure*}
    \centering
    \includegraphics[width=6in]{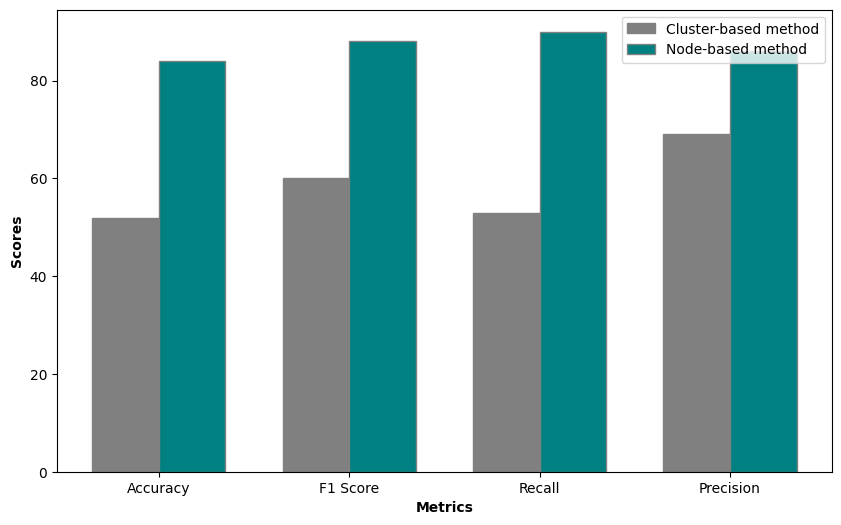}
    \caption{Node-based comparison on Medical diagnosis use-case.}
    \label{fig:ds}
\end{figure*}

In Figure 5, the metrics constructed using True Positive (TP), False Positive (FP), True Negative (TN), and False Negative (FN) are presented. Our analysis focused on evaluating accuracy, F1 score, recall, and precision based on these metrics. The comparison depicted in the figure highlights the performance disparity between the node-based and cluster-based methods. Notably, the node-based method emerges as the clear frontrunner, exhibiting significantly higher accuracy, F1 score, recall, and precision compared to the cluster-based approach. This observation underscores the effectiveness of the node-based method in accurately capturing connections within the knowledge graph, ultimately leading to superior performance across all evaluated metrics.

The observed limitations in accuracy, F1 score, recall, and precision of the cluster-based method can be attributed to the utilization of Clinical BERT. While Clinical BERT is a powerful pre-trained model, its efficacy is constrained by practical considerations such as computational resources and sample size limitations. Due to the vast scale of clusters within the knowledge graph, it becomes necessary to divide them into smaller, manageable chunks for processing. However, this segmentation introduces a challenge: the loss of contextual coherence across multiple chunks. As a consequence, the embeddings derived from fragmented clusters may lack the holistic context necessary for accurate representation and inference, resulting in diminished performance metrics. This phenomenon underscores the importance of considering both the capabilities and limitations of pre-trained models when designing and implementing knowledge graph construction methodologies.

In addition to evaluation measures, transparency and time are critical factors in assessing the effectiveness of the M-KGA approach. As depicted in Figures 3 and 4, transparency refers to the clarity and comprehensibility of the constructed knowledge graph. The node-based method excels in transparency by establishing direct connections between nodes, thereby presenting a clear and intuitive representation of relationships. In contrast, the cluster-based approach may exhibit less transparency, as it tends to add fewer relationships, resulting in a less explicit depiction of connections.

Time, on the other hand, pertains to the efficiency of the knowledge graph construction process. The cluster-based approach demonstrates an advantage in terms of time efficiency, requiring less computational resources and processing time compared to the node-based method. However, this efficiency comes at a cost, as the cluster-based approach may sacrifice performance metrics such as accuracy, F1 score, recall, and precision, as previously discussed.

Overall, while the cluster-based approach offers a quicker construction process, it may compromise transparency and performance. In contrast, the node-based method prioritizes transparency and performance, albeit at the expense of increased computational complexity and time consumption. Thus, the choice between these approaches should be carefully considered based on the specific requirements and priorities of the knowledge graph application.

\section{Conclusion and Future Work}

In conclusion, this study introduces the Medical Knowledge Graph Automation (M-KGA) approach, which aims to address the challenges associated with automating the construction of knowledge graphs (KGs) and enhancing their completeness. Leveraging user-provided medical concepts and BioPortal ontologies, M-KGA enriches the semantic content of KGs using pre-trained embeddings, thereby facilitating a more comprehensive representation of structured medical knowledge. Our approach incorporates two distinct methodologies, namely a cluster-based approach and a node-based approach, to uncover hidden connections within the knowledge graph.

Through rigorous testing involving 100 medical concepts, our M-KGA framework demonstrates promising results, showcasing its potential to overcome the limitations of existing knowledge graph automation techniques. The performance metrics and graph visualizations presented in this study underscore the effectiveness of our approach in enhancing the transparency and accuracy of knowledge graphs, particularly in the medical domain.

Looking ahead, future work will focus on addressing scalability issues associated with the cluster-based method, aiming to improve its performance. Additionally, we plan to explore retrieval augmented generation (RAG) with Large Language Models (LLMs), for knowledge graph development and performance comparison with our current approach. By continuing to innovate and refine our approach, we aim to further advance the field of knowledge graph automation and contribute to the development of more comprehensive and accurate representations of structured knowledge in healthcare domain.

\bibliography{sample}

\end{document}